\documentclass[letterpaper, 10 pt, conference]{ieeeconf}  
\IEEEoverridecommandlockouts                              
\usepackage{graphicx}
\usepackage{amsmath}
\title{\LARGE \bf Learning to Push by Grasping: Using multiple tasks for effective learning}
\author{Lerrel Pinto and Abhinav Gupta\\
Carnegie Mellon University
}
\begin{document}
\maketitle
\thispagestyle{empty}
\pagestyle{empty}
\begin{abstract}
Recently, end-to-end learning frameworks are gaining prevalence in the field of robot control. These frameworks input states/images and directly predict the torques or the action parameters. However, these approaches are often  critiqued  due  to  their huge  data  requirements for learning a task. The argument of the difficulty in scalability to multiple tasks is well founded, since training these tasks often require hundreds or thousands of examples. But do  end-to-end  approaches  need  to  learn  a unique  model  for  every  task?  Intuitively, it seems that sharing across tasks should help since all tasks require some common understanding of the environment. In  this  paper,  we  attempt  to  take  the  next  step  in  data-driven end-to-end learning frameworks: move from the realm of  task-specific  models  to  joint  learning  of  multiple  robot tasks. In an astonishing result we show that models with  multi-task  learning  tend  to  perform  better  than  task-specific  models  trained  with  same  amounts  of  data.  For example,  a  deep-network  learned  with  2.5K grasp and  2.5K  push  examples performs better on grasping than a network trained on 5K  grasp  examples. 
\end{abstract}

\section{INTRODUCTION}
Consider a robot trying to manipulate (e.g., grasp or push) an object. To perform successful grasps, a robot would need to (a) infer object properties (geometry, mass distribution etc.); (b) have a knowledge of its own anatomy and (c) finally understand what makes a successful grasp and how to achieve it. Analytical frameworks such as ~\cite{brooks1983planning,shimoga1996robot,miller2004graspit} focus on defining (c) via mathematical framework. These approaches assume object properties (such as 3D geometry) are either given or estimated by a separate perception pipeline. They also make several simplifying assumptions such as uniform density distributions, simplified friction models etc. Due to strong reliance on perception pipelines, these approaches have not shown promising results. Alternatively, end-to-end learning approaches have been gaining prominence~\cite{pinto2015supersizing,levine2016learning,pinto2016curious,agrawal2016learning}. These approaches combine (a)-(c) and learn a joint model in a data-driven manner. Specifically, they collect thousands of examples of successful and unsuccessful manipulations~\cite{pinto2015supersizing,levine2016learning,agrawal2016learning} and then learn a model which controls the manipulation directly from input images.

While end-to-end learning frameworks have been quite promising, they are often viewed skeptically due to their huge data requirements. Critics often argue that most end-to-end models require training a unique model for every task and since each model require thousands of examples, the approach is not scalable. But do end-to-end approaches need to learn unique models? Is there some kind of sharing across tasks that can boil down data-requirements? Intuitively, it seems like sharing should help: all tasks require perception of object properties (a) and learning robot's parameters (b). Therefore, at the very least, data collected for say pushing objects should be useful in learning perception modules for grasping objects as well. 

In this paper, we attempt to take the next step in data-driven end-to-end learning frameworks:  move from the realm of task-specific models to joint learning of multiple robot tasks. In an astonishing result we show that models with multi-task learning tend to perform better than task-specific models trained with same amounts of data. For example, a deep-network learned with 5K grasp examples tends to work worse than a network trained with 2.5K grasps and 2.5K push examples. We hypothesize that 
performing alternate tasks may expose object properties and modalities that are inaccessible to the original task. Both grasping and pushing are dependent on object properties like geometry. However pushing an object may reveal modalities of the object's properties different from grasping that object. The joint learning of these multiple tasks also acts as a regularization leading to learning of more generalizable features. This would therefore improve performance on previously unseen objects.

\begin{figure}[t!]
\begin{center}
\includegraphics[width=3.3in]{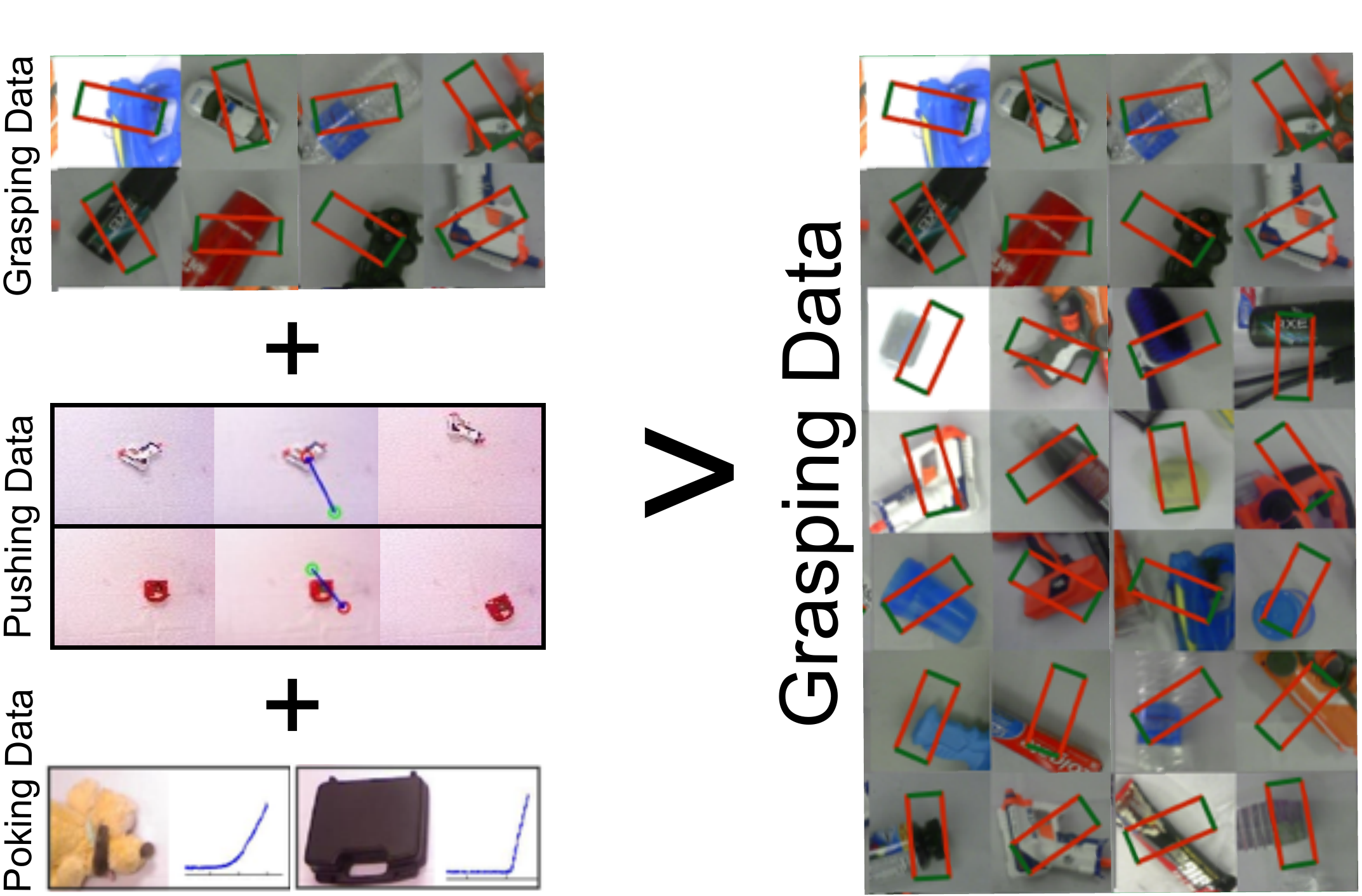}
\end{center}
\caption{By using a multi task learning framework, we learn shared representations of tasks that improve the performance on individual tasks as well. We show that the benefit of additional data from different tasks is often more important than data from the original task.}
\label{fig:intro_fig}
\end{figure}


\section{RELATED WORK}

\noindent {\bf Grasping:} Grasping is one of the oldest open problems in the field  of robotics.  For a  comprehensive  literature  review, we direct the readers to ~\cite{bicchi2000robotic,bohg2014data}. Most of the initial work in this field focused on analytical methods and 3D reasoning for predicting grasp locations and configurations~\cite{brooks1983planning,shimoga1996robot,lozano1989task,nguyen1988constructing}. The Graspit \cite{miller2004graspit,miller2003automatic} software then allowed for simulations to rank grasp candidates. More recently, data-driven learning-based approaches have started to appear, with initial work focused on using human annotators~\cite{lenz2013deep}. This was followed by our self-supervized learning framework ~\cite{pinto2015supersizing} where we used a robot to continuously collect grasp data. Google's work ~\cite{levine2016learning} then scaled this up by using multiple robots collecting data in parallel.

\noindent {\bf Pushing:} Pushing is another fundamental robotics task. The origins of pushing as a manipulation task can be traced to the task of aligning objects to reduce pose uncertainty~\cite{balorda1990reducing,balorda1993automatic,lynch1996stable} and as a preceding realignment step before object manipulation ~\cite{dogar2011framework,yun1993object,zhou2016convex}. Pushing also offers a method of moving objects without needing to explicitly grasp them ~\cite{katz2014perceiving}. Recently there has been a lot if interest in using pushing to learn intuitive physics ~\cite{fragkiadaki2015learning, mottaghi2016happens} on simulators. ~\cite{pinto2016curious, agrawal2016learning} collect robot executed pushing data to further analyse the effects of push manipulation. ~\cite{agrawal2016learning} also show how deep learning on 50K push data can be used to push objects in the real world. 

\noindent {\bf Tactile Sensing (Poking):} We use tactile response prediction as an auxiliary task to help learning how to push and grasp. The utility of learning tactile responses for visual representations has been shown in  ~\cite{pinto2016curious,schneider2009object}.

\noindent {\bf Multi task learning:} The recent work mentioned show how collecting large amounts of data can be used to learn and perform robot tasks. However, a major concern with these self-supervized frameworks is the collection of data, which is time intensive. Multi task learning (MTL) is generally used to model related tasks closely ~\cite{misra2016cross, gkioxari2014r, kokkinos2016ubernet} by using a shared representation and exploiting the commonality and structure of these tasks. Often sequential MTL or `finetuning' helps train deep network models on tasks that have very limited data ~\cite{ren2015faster}, by initializing the parameters from a previous task. What we are interested in is joint MTL where both the tasks are simultaneously learnt.

Our work build upon our past work (see, ``The Curious Robot'')  which trains visual representation~\cite{pinto2016curious} using multiple tasks including grasping, pushing, and poking. However, the previous work  \cite{pinto2016curious} demonstrated how physical robot tasks can be useful for non-physical tasks such as image retrieval. No performance on original training tasks was reported. In this paper, we focus on developing an end-to-end multi-task learning of physical tasks such as grasping and pushing. 

When it comes to robotic tasks, most frameworks focus on task-specific models and learning. To the best of our knowledge, this is one of the first efforts that report sharing across physical tasks can help improve performance on these tasks. In this work, we show that MTL not only helps improve the performance of both grasping and pushing tasks, but shows that the value of an additional datapoint of the original task is less than that of the alternate task. We believe that the sharing of representations enable robust and regularized feature learning that helps in improving both tasks. Hence, we exploit MTL and show that with even less amounts of data, efficient models can be learnt by using big data from other tasks.


\section{OVERVIEW}
Our goal is to explore if data collected for one task such as grasping can be helpful in training representations and control for other tasks such as pushing or poking.  Based on current research trends, it seems current work focuses on training representations and control models specific to a task. However, we argue that most of these tasks face a common challenge of learning how the world works and therefore can share the data to learn faster. Specifically, our core hypothesis is that some of the parameters in the Convolutional Neural Network (ConvNet) correspond to learning visual features. A few parameters correspond to learning the underlying structure and physics. There should parameters corresponding to the anatomy and general control information of the robot. Finally, the remaining parameters should be specific to every task that is being learned. If this is indeed the case, data sharing across tasks could be vital for learning the parameters for visual representation, structure/physics and the robot-specific control. 

In this paper, we investigate if multi-task learning can help learn a better control model for the tasks of grasping and pushing. We collect data using three tasks: {\bf (a) grasping:} the robot attempts to grasps the objects on table-top setting; the sensor in the gripper measures the success/failure on the task; {\bf (b) pushing:} the robot pushes the objects on the table with specific force and observes the initial and final states to learn a mapping between actions and state transformations; {\bf (c) poking:} the robot uses a skin-sensor on the finger to push objects into the table and observe the force in the sensor as object is pushed. Finally, we explore how a Grasp ConvNet trained using grasp data alone performs in comparison to a Grasp ConvNet trained using grasping, pushing and poking data. We also compare how a Push ConvNet trained using push data alone performs in comparison to Push ConvNet trained using grasping, pushing and poking.

\section{APPROACH}
We now describe the formulation of our manipulation tasks: planar grasping and planar pushing. We also describe the poking data and how it is incorporated into the framework.

\subsection{Planar Grasps} 
\begin{figure*}[t!]
\begin{center}
\includegraphics[width=6.0in]{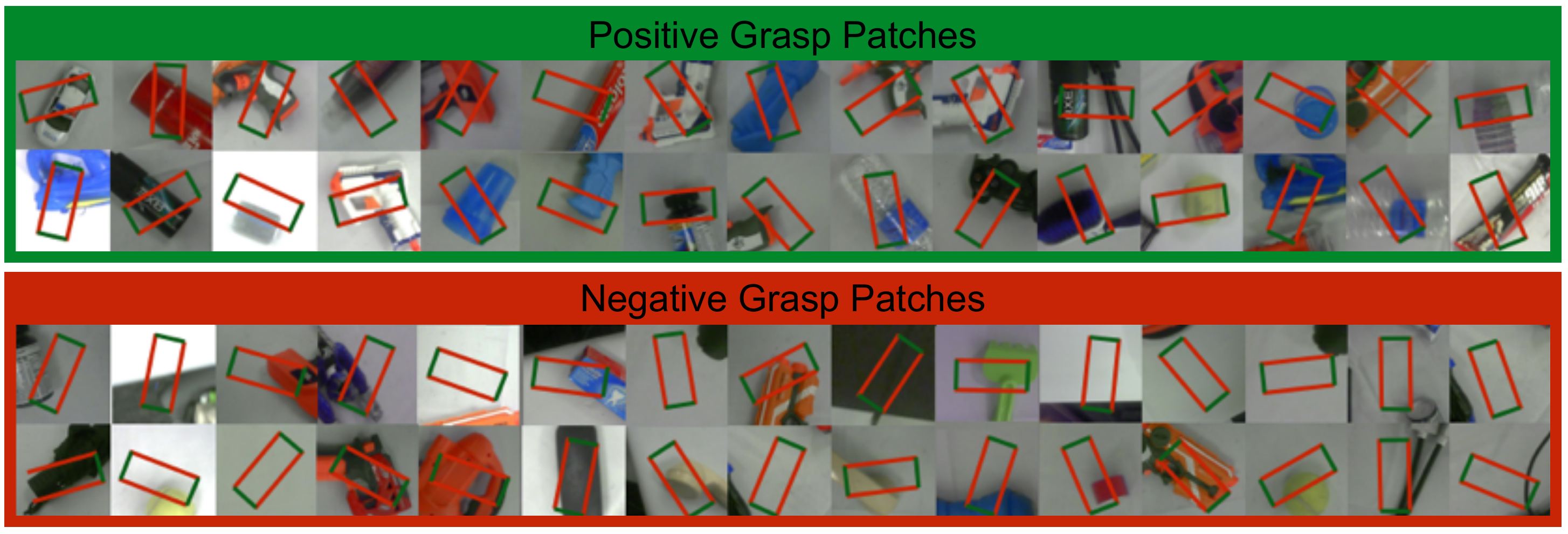}
\end{center}
\caption{Examples of successful (top) and unsuccessful grasps (bottom). This data is taken from ~\cite{pinto2015supersizing}. We use a patch based representation: given an input patch we predict 18-dim vector which represents whether the center location of the patch is graspable at $0^{\circ}$, $10^{\circ}$, \dots $170^{\circ}$.}
\label{fig:grasp_data}
\end{figure*}

We use the grasp dataset described in our earlier work \cite{pinto2015supersizing} for our experiments on the grasping task. The grasp configuration can be defined using 3 parameters, $(x,y,\theta)$: position of grasp point on the surface of table and angle of grasp. The training dataset contains around 37K failed grasp interactions and around 3K successful grasp interactions as the training set. For testing, we use around 2.8K failed and 0.2K successful grasps on novel objects are provided. We use this training set to evaluate the performance of our network. Some of the positive and negative grasp examples are shown in Figure~\ref{fig:grasp_data}.

\vspace{2pt}
\noindent\textit{\bf Grasp prediction formulation:} 
The grasp prediction problem can be formulated as finding a successful grasp configuration $(x_S,y_S,\theta_S)$ given an image of an object $I$. However, as mentioned in \cite{pinto2015supersizing,pinto2016curious}, this formulation is problematic due to the presence of multiple grasp locations for each object. Hence to encode the $(x_S,y_S)$ configuration, we sample a patch $I_G$ centered at the location of $(x_S,y_S)$ in the image $I$.  Given an image patch, we output an 18-dimensional likelihood vector where each dimension represents the likelihood of whether the center of the patch is graspable at $0^{\circ}$, $10^{\circ}$, \dots $170^{\circ}$. Therefore, the grasping problem can be thought of as 18 binary classification problems. Hence the evaluation criterion is binary classification i.e. given a patch and executed grasp angle in the test set, predict whether the object was grasped or not.

\subsection{Planar Push}
We use the push data collected in our previous work ~\cite{pinto2016curious}, in which a Baxter robot collects push data. The dataset contains images of objects before and after a push is acted on the object. Each data-point consists of the initial image $I_{begin}$, the push action $A_P  = (x_{start},y_{start},x_{final},y_{final},z_{pushHeight})$ and the final image $I_{end}$.

\begin{figure*}[t!]
\begin{center}
\includegraphics[width=7.0in]{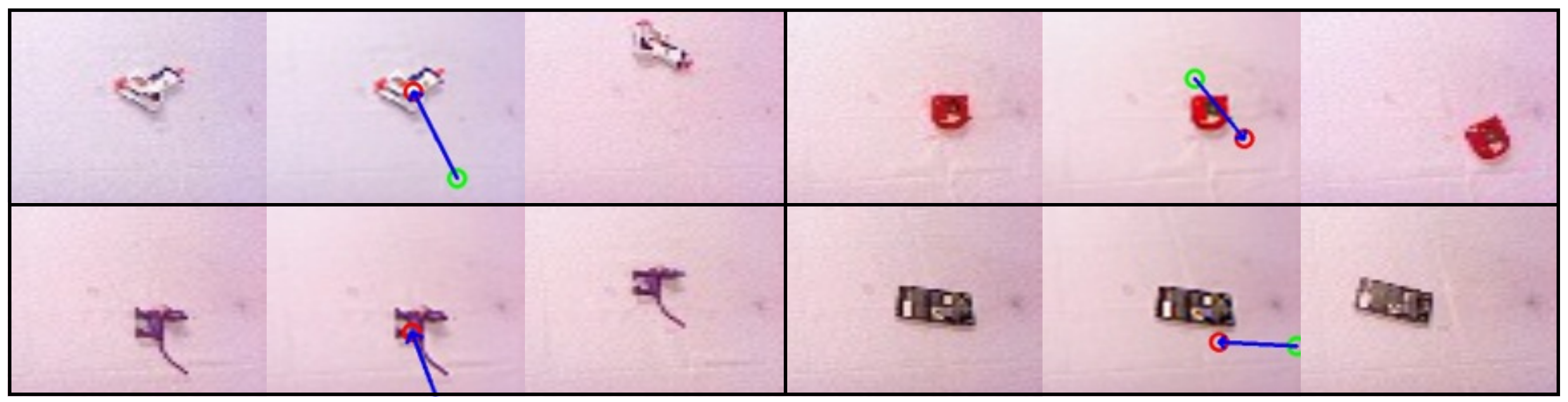}
\end{center}
\caption{Examples of push action on 4 objects. For each object, left image shows the image of object before the push, middle shows the push action on the object and the right image shows the resultant image after the push. The linear push action is described as an arrow with the green circle being the start configuration for the end effector and the red circle being the final configuration of the end effector.}
\label{fig:push_data}
\end{figure*}

The dataset contains 5K push actions on 70 objects using the above described method. Some of these push actions are visualized in Figure \ref{fig:push_data}. 

\begin{figure}[t!]
\begin{center}
\includegraphics[width=3.3in]{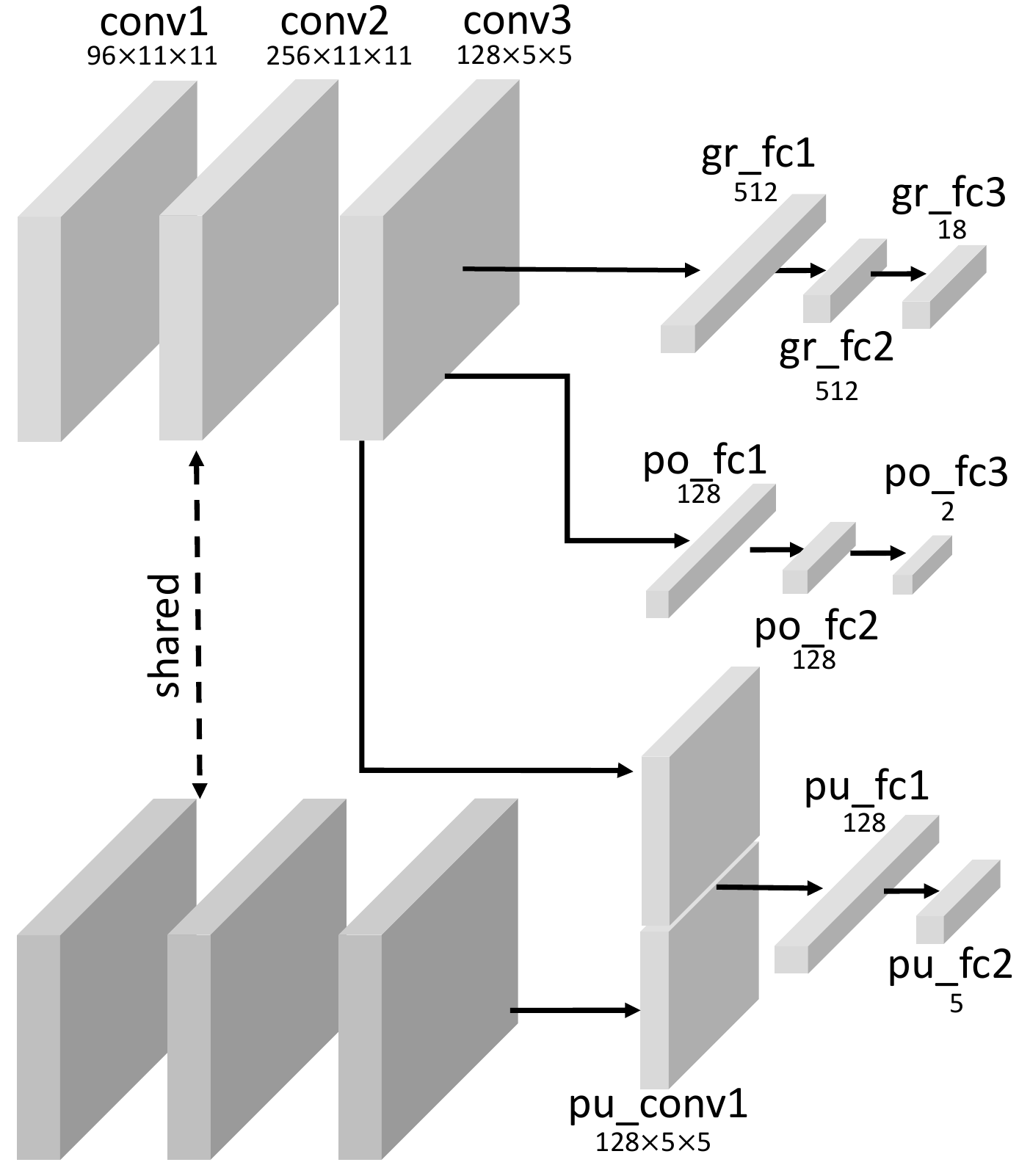}
\end{center}
\caption{The shared network for training grasp, poke and push simultaneously allows for learning robust and generalizable lower level features that help the learning of the inidividual tasks as well. The first 3 conv layers are shared which is followed by task specific layers for the individual tasks.}
\label{fig:multitask_network}
\end{figure}

\noindent\textit{\bf Push prediction formulation:} The task for the learner is to now predict the push action $A_P$ given the images $I_{begin}$ and $I_{end}$. To learn this, we use a siamese network with shared weights. One tower of this network takes $I_{begin}$ as input and the second tower takes $I_{end}$ as input. The siamese outputs are then concatenated and followed by fully connected layers to regress to the push action that caused this transformation. The loss function for regression is the euclidean loss. Note that this action formulation captures the relevant magnitude as well as the localization and direction of the push.

\subsection{Planar Poke}
\begin{figure*}[t!]
\begin{center}
\includegraphics[width=6.5in]{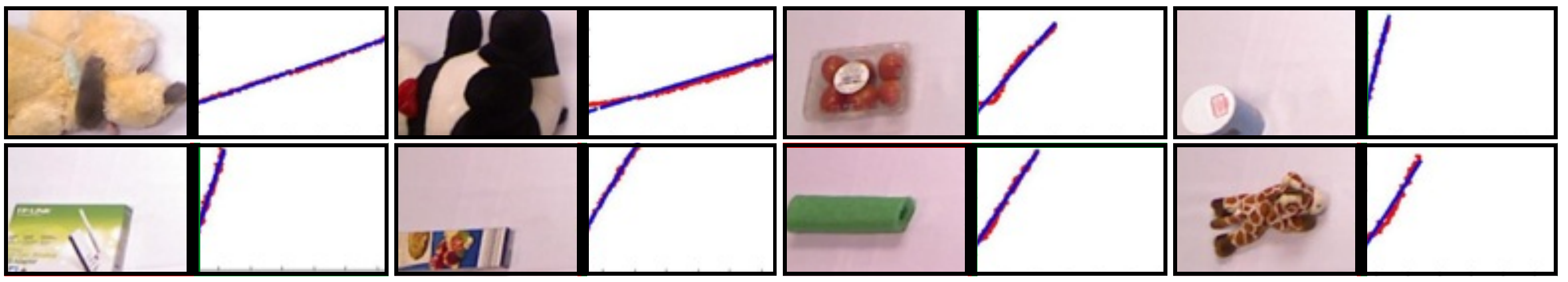}
\end{center}
\caption{Examples of poking on 8 objects from the dataset in ~\cite{pinto2016curious}. For each object, left images shows the image of object before the poke, the right images show the tactile force response during the push. A line is fitted on this response (blue line) which is used for final regression}.
\label{fig:poke_data}
\end{figure*}

We use the tactile poke data collected in our previous work ~\cite{pinto2016curious}. In that work, a Baxter robot collected poke data by pushing objects into a table. The dataset contains images of objects and the tactile force felt while poking the object. Each datapoint consists of the image $I_{poke}$ and the poke response $R_P$. The task for the learner is to now predict the poke response $R_P$ given the image of the object $I_{poke}$. Examples of this data can be see in Figure ~\ref{fig:poke_data}.

\noindent\textit{\bf Poke prediction formulation:} We parametrize the poke response function $R_P$ with 2 parameters (slope and intercept of voltage increase). Therefore, an auxiliary task for the learner is to predict the poke response $R_P$ given the image of the object $I_{poke}$. To do this we use a very similar network as the grasping network, where the first 3 layers are shared and the rest are learnt for the poking task independently. The loss function here is euclidean unlike the binary loss in grasping. The last layer of network has 2 neurons, corresponding to the linear parametrisation of the tactile response.

\subsection{Network Architecture}
We now describe the architecture of our multi task network in Figure ~\ref{fig:multitask_network}. Since we require the transfer of features learnt from (a) grasping, poking to pushing and (b) pushing, poking to grasping, we use a common shared network for the lower layers. These shared layers are then connected to task specific layers for grasping, pushing and poking. 

\subsubsection{Input to the network}
The input to the network is a $64\times64\times3$ image. Hence all the training and testing images are resized to $64\times64$.

\subsubsection{Shared representation}
The first three convolutional layers are shared between the two tasks. The first convolutional layer (conv1) has 96 kernels with $11\times11$ kernel size. This is followed by a batch normalization (BN) layer ~\cite{ioffe2015batch} and ReLU ~\cite{krizhevsky2012imagenet} as the non linearity. The second convolutional layer (conv2) has 256 kernels with $11\times11$ kernel size. Once again the outputs from the conv2 go through a BN and ReLU. The third convolutional layer (conv3) has 128 kernels with size $5\times5$ followed by a BN and ReLU.

\subsubsection{Grasp specific network} 
For the grasping task, the output from the shared layers is input to a fully connected layer (gr\_fc1) with 512 neurons. This is followed by a dropout layer ~\cite{krizhevsky2012imagenet} with drop probability 0.5 and a ReLU. The second fully connected layer (gr\_fc2) contains 512 neurons and is followed by another dropout with drop probability 0.5 and ReLU. The final fully connected layer (gr\_fc3) has 18 neurons which correspond to the 18 angles of grasp we are trying to classify.

\subsubsection{Push specific network}
Note that the input for the push prediction problem is 2 images ($I_{begin}$ and $I_{end}$). Hence to use the shared representation, we have a siamese architecture with a replicated network. $I_{begin}$ goes through one tower of the siamese while $I_{end}$ goes through the second tower. The conv3 representations of these two images go through a further convolutional layer (pu\_conv1) with 128 kernels of $5\times5$ kernel size. The two pu\_conv1 representations are then concatenated and is followed by a fully connected layer (pu\_fc1) with 128 neurons. This is followed by a dropout layer with drop probability 0.5 and a ReLU. The final fully connected layer (pu\_fc2) contains 5 neurons that correspond to the 5 dimensional action $P_A$ that we are trying to regress to.

\subsubsection{Poke auxiliary network}
The input for the poke prediction problem is the image $I_{poke}$. We again use the shared representation for the first 3 conv layers followed by 3 fully connected layers. These 3 fully connected layers (po\_fc1, po\_fc2, po\_fc3) are similar to the one in grasping network and has 128, 128 and 2 neurons in each of the layers respectively.

\subsection{Learning:}
We now describe details of learning the multi task network parameters along with the loss functions for the individual tasks. Let us denote the learnable parameters in the shared representation (conv1, conv2 and conv3) as $W_S$, the parameters in grasp specific network (gr\_fc1, gr\_fc2 and gr\_fc3) as $W_G$ and the parameters in push specific network (pu\_conv1, pu\_fc1, pu\_fc2) as $W_P$. Let the grasp network be denoted by the function $G$, the push network denoted by the function $P$ and the poke network denoted by the function $Poke$.

\subsubsection{Grasp loss}
Given an input patch $I_G$, the attempted discrete grasp angle $\theta_D\in\{0,1,2,..17\}$ and the grasp success label $y\in\{0,1\}$, let the predicted success be $y^\prime=G(I_G;W_S,W_G)[\theta_D]$. This denotes the $\theta^{th}_D$ neuron output of the gr\_fc3 layer. The binary cross entropy loss for this element $L_G$ is:
\begin{equation}
L_G = -y\times log(sig(y^\prime)) - (1 - y) \times log(1 - sig(y^\prime))
\end{equation}

\noindent Here $sig$ is the sigmoid function $sig(y^\prime) = 1/(1+e^{-y^\prime})$. The definition of this loss allows the network to learn multimodal distributions of grasp angles since the loss is only dependent on the attempted angle.

\subsubsection{Push loss}
Given the input patches $I_{begin}$ and $I_{end}$, and the executed push action $p_A$, let the predicted push action be $p^{\prime}_A=P(I_{begin},I_{end};W_S,W_P)$. The euclidean loss for this element $L_P$ is:
\begin{equation}
L_P = (p_A - p^{\prime}_A)^2
\end{equation}

\subsubsection{Poke loss}
Given the input image $I_{poke}$, and the poke response $p_R$, let the predicted poke response be $p^{\prime}_R=Poke(I_{poke};W_S,W_{Poke})$. The euclidean loss for this element $L_{Poke}$ is:
\begin{equation}
L_{Poke} = (p_R - p^{\prime}_R)^2
\end{equation}

\begin{figure*}[t!]
\begin{center}
\includegraphics[width=7in]{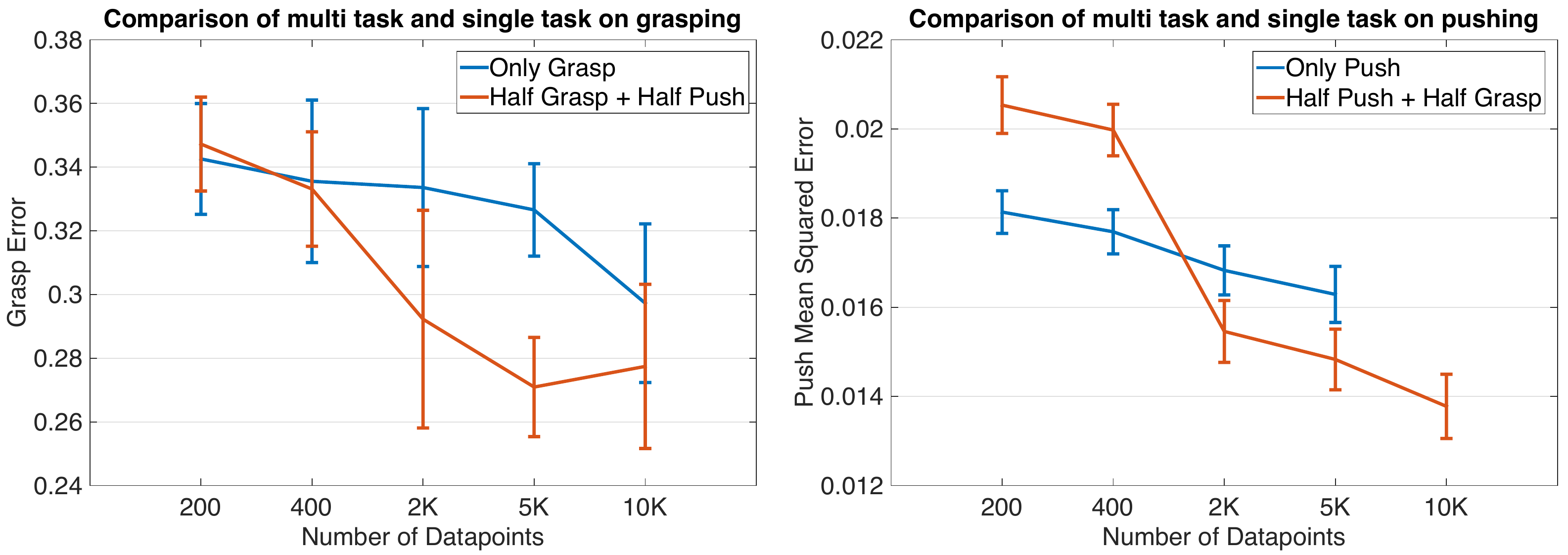}
\end{center}
\caption{We show the extent to which replacing half data of the original task with half data for of a different task helps the original task. For the grasping task on the left, grasp error is essentially 1-grasp accuracy on a novel object test set. While for the pushing task, the error is mean squared error on predictions over a novel object test set. We can see that for both the grasping and the pushing task, dropping 50\% of the data and replacing it with the other task data helps in some data size regimes. This gain is quite significant for the pushing task.}
\label{fig:50comparison}
\end{figure*}

\subsubsection{Joint training}
For one iteration of training, given the training set for the grasp and pushing task, a batch of $n=128$ elements is randomly selected. This gives us a grasp data batch $B_G$, a push data batch $B_P$ and a a poke data batch $B_{Poke}$. Note that each of the 128 elements of the grasp data batch contains $I_G$, $\theta_D$ and $y^\prime$. Similarly each element of the push data batch contain a $I_{begin}$, $I_{end}$, and $p_A$ and that of the poke data batch contain $I_{poke}$ and $p_R$.

\noindent The cumulative loss for the grasp batch is $L_BG =\frac{1}{n} \sum\limits_{i=1}^nL_{Gi}$, where $L_{Gi}$ is the loss from grasp batch element $i\in\{1:n\}$. Similarly the cumulative loss for the push batch is $L_BP =\frac{1}{n}\sum\limits_{i=1}^nL_{Pi}$, where $L_{Pi}$ is the loss from the push batch element $i\in\{1:n\}$. The loss for the poke batch is $L_BPoke =\frac{1}{n}\sum\limits_{i=1}^nL_{Pokei}$, where  where $L_{Pokei}$ is the loss from the poke batch element $i\in\{1:n\}$

\noindent During training, first the gradients $\frac{\partial L_BG}{\partial W_S}$ and $\frac{\partial L_BG}{\partial W_G}$ are computed from the grasp data batch. The gradients $\frac{\partial L_BP}{\partial W_S}$ and $\frac{\partial L_BP}{\partial W_P}$ are then computed from the push data batch, gradients $\frac{\partial L_BPoke}{\partial W_S}$ and $\frac{\partial L_BPoke}{\partial W_{Poke}}$ are then computed from the poke data batch . Note that the total loss gradient with respect to $W_S$ is $\frac{\partial (L_BG + L_BP + L_BPoke)}{\partial W_S}$.

\noindent The parameters are then updated as:
\begin{align}
&W_S \leftarrow \texttt{RMSProp(}W_S,\frac{\partial (L_BG + L_BP +L_BPoke)}{\partial W_S}\texttt{)}\nonumber\\
&W_G \leftarrow \texttt{RMSProp(}W_G,\frac{\partial L_BG}{\partial W_G}\texttt{)}\nonumber\\
&W_P \leftarrow \texttt{RMSProp(}W_P,\frac{\partial L_BP}{\partial W_P}\texttt{)}\nonumber\\
&W_{Poke} \leftarrow \texttt{RMSProp(}W_{Poke},\frac{\partial L_BPoke}{\partial W_{Poke}}\texttt{)}\nonumber
\end{align}

\noindent Here RMSProp is a gradient descent approach ~\cite{tieleman2012lecture}. We use a learning rate of 0.002, momentum of 0.9 and decay of 0.9. The learning rate decays at a schedule of 0.1 factor every 5000 iterations.

\section{RESULTS}
We now describe our results for multi-task learning as compared to learning for a specific task using that task data alone. For quantitative evaluations, we only use grasping and pushing task.  For all our experiments, we use multiple folds of data to make sure just one set of data is not showing the behaviour. 

{\bf Error Metrics:} For evaluating the grasping, we use classification error. For an input patch and angle, the grasping network has to classify whether the grasp will be successful or not. For evaluating the push prediction, we use mean squared error as the metric.

\subsection{Evaluating Multi-Task vs. Task-Specific} As our first experiment, we evaluate the performance of multi-task framework as compared to task-specific training. We only use the multi-task network for pushing and grasping (ignoring the poke task). We compare the performance keeping the total number of training data points as constant. For multi-task training, we use 50\% data from pushing and 50\% data from grasping. Figure~\ref{fig:50comparison} shows the comparison with respect to the total amount of training data.

The results are surprising. When the total amount of training data exceeds thousand examples, multi-task seems to outperform task-specific network. This seems to suggest that in this regime, training datapoint for a different task is more important than the training datapoint of the original task. Our hypothesis is that the multi-task datapoints provide diversity in training and it also provides regularization due to extra loss function (preventing overfitting). An interesting observation is that for both tasks in low data regime, task specific network works better. This is primarily because we need a minimum number of datapoints for training task-specific layers in the network.

\begin{figure*}[t!]
\begin{center}
\includegraphics[width=7in]{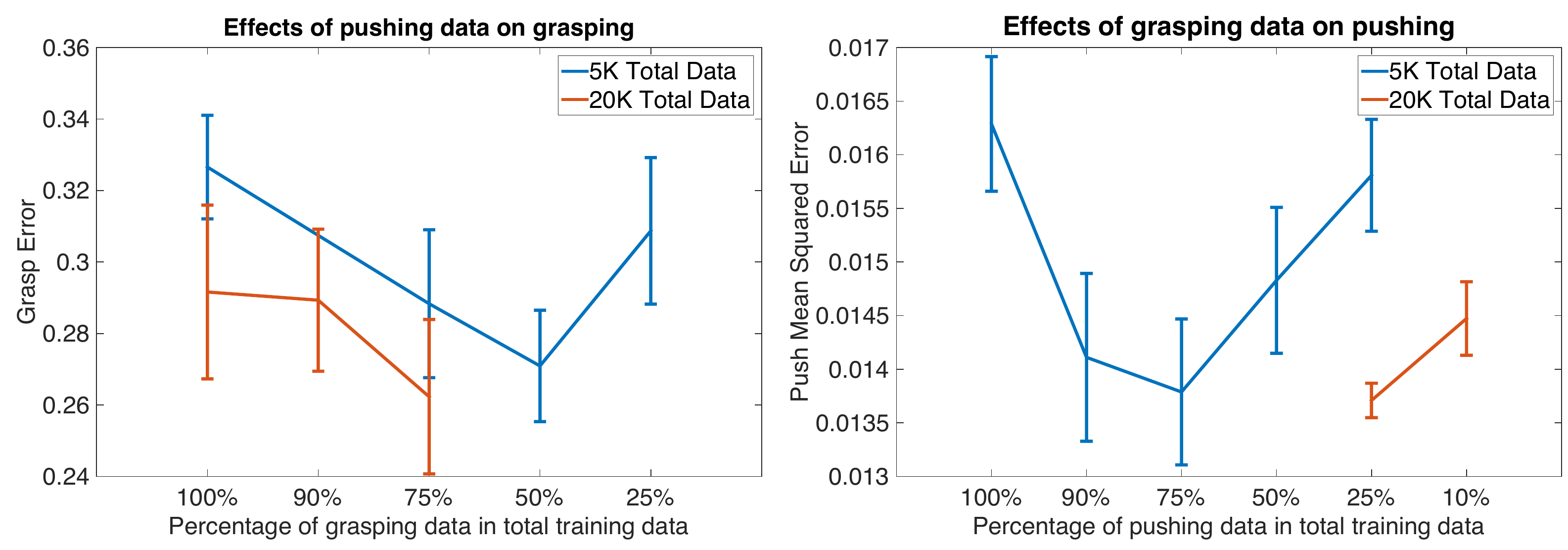}
\end{center}
\caption{We study the different data replacement percentages on the performance of the individual tasks. Given 5K training data, for the grasping task, replacing 50\% of the grasping data with pushing task data, helps the most. While for the pushing task, replacing 75\% of the pushing data with data for grasping task helps the most.}
\label{fig:push_grasp_comp}
\end{figure*}

\subsection{Multitask: Data Ratio}
Next, we want to evaluate what data ratio is better for training the multi-task framework. For example, for training a multi-task network for grasping: is it better to have 50\% training samples each from grasping and pushing or is it better to have 75\% grasping and 25\% pushing. Specifically, we varied the ratio ($r$) of number of datapoints for original task divided by number of total datapoints. Figure~\ref{fig:push_grasp_comp} shows the performance for both the tasks of grasping and pushing. We use 5K and 20K as total number of samples.

It can be seen from the figure that at 50\% of training data grasping has the best performance. On the other hand for the pushing task, the best ratio seems to be 75\% of training data has to be pushing. It seems pushing is able to transfer more knowledge to grasping as compared to the reverse.

\subsection{Multitask: 3-task performance}
Finally, we want to evaluate the multi-task performance if three tasks are used. Specifically, we evaluate if adding poking as another task can improve the performance better. We also evaluate how changing the ratios of different data would behave on grasping and pushing.

\begin{figure*}[t!]
\begin{center}
\includegraphics[width=7in]{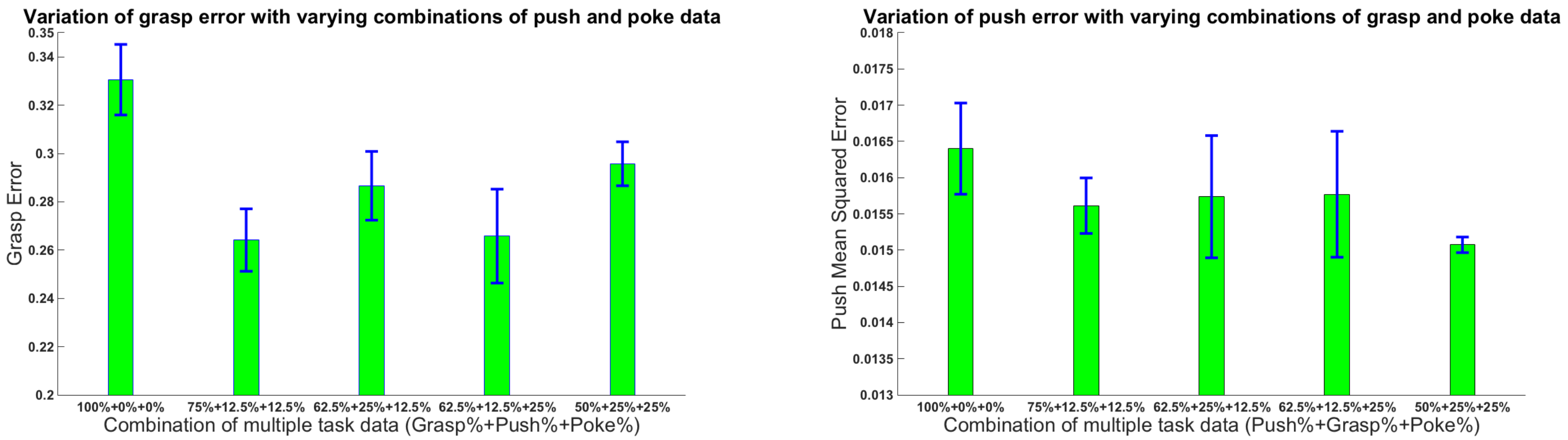}
\end{center}
\caption{We now show the results on jointly training the 3 tasks. Given different proportions of pushing and poking data, we see improvements in the performance of grasping. Similarly for the pushing task, we present results with varying proportions of grasping and poking data. }
\label{fig:poke_comp}
\end{figure*}

Figure~\ref{fig:poke_comp} shows the performance on the grasping and pushing task. Note that the best performance we get for total 4K data on grasping is 28\% when trained using two tasks. However, using 3 tasks the best performance has a reduced error rate of only 26\%. This demonstrates that poking can also help improve the performance of the task. It also seems for grasping, poking
data is more important than the pushing data (see comparison of grasp error between 62.5\%Grasp+25\%Push+12.5\%Poke vs 62.5\%Grasp+25\%Poke+12.5\%Push). 

\subsection{Qualitative Results}
\begin{figure*}[t]
\begin{center}
\includegraphics[width=7in]{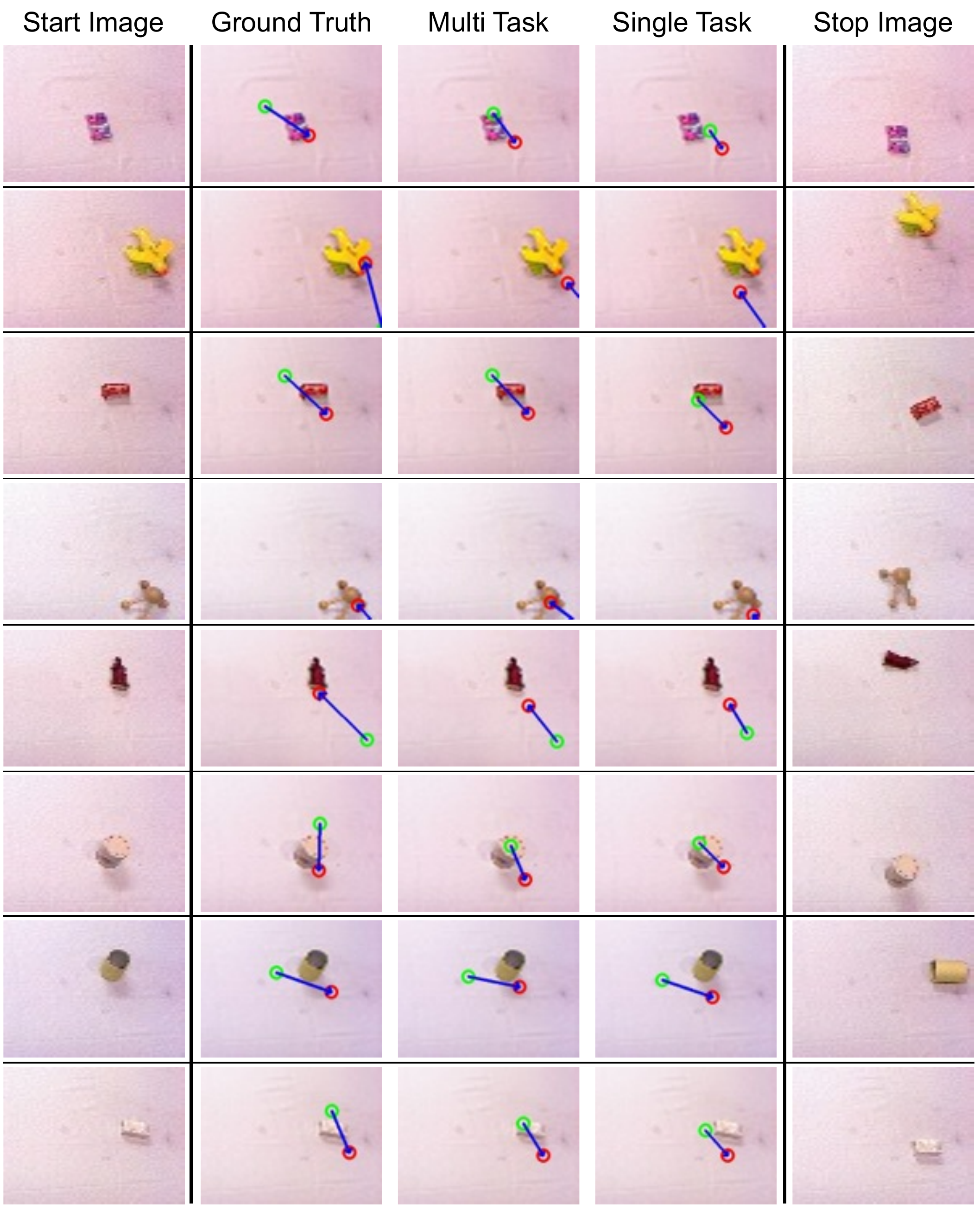}
\end{center}
\caption{We show the qualitative results for the pushing task, randomly sampled from the test set. The green circle represents the start configuration for the end effector, the red circle represents the final configuration of the end effector and the blue arrow represents the direction of push.}
\label{fig:qual_fig}
\end{figure*}

Figure~\ref{fig:qual_fig} shows the qualitative results on pushing. As it can be seen from the figure multi-task network performs significantly better.

\section{DISCUSSION}
Most of the current research in robotic control focuses on learning for specific tasks. The general consensus in the community seems to be that sharing across tasks does not really help. This exacerbates the problem and presents an issue for end-to-end learning approaches since it would mean large amounts of data needs to be collected for every task. This paper attempts to break the myth of task-specific learning and shows that multi-task learning is not only effective but in fact improves the performance even when the total amount of data is the same. We hypothesize this is primarily because of diversity of data and regularization in learning. This paper opens up a new subfield of multi-task learning in robotics; specifically focusing on mechanism of sharing across different tasks.

\bibliographystyle{unsrt}
\bibliography{references} 

\begin{thebibliography}{10}

\bibitem{brooks1983planning}
Rodney~A Brooks.
\newblock Planning collision-free motions for pick-and-place operations.
\newblock {\em IJRR}, 2(4):19--44, 1983.

\bibitem{shimoga1996robot}
Karun~B Shimoga.
\newblock Robot grasp synthesis algorithms: A survey.
\newblock {\em IJRR}, 15(3):230--266, 1996.

\bibitem{miller2004graspit}
Andrew~T Miller and Peter~K Allen.
\newblock Graspit! a versatile simulator for robotic grasping.
\newblock {\em Robotics \& Automation Magazine, IEEE}, 11(4):110--122, 2004.

\bibitem{pinto2015supersizing}
Lerrel Pinto and Abhinav Gupta.
\newblock Supersizing self-supervision: Learning to grasp from 50k tries and
  700 robot hours.
\newblock {\em ICRA}, 2016.

\bibitem{levine2016learning}
Sergey Levine, Peter Pastor, Alex Krizhevsky, and Deirdre Quillen.
\newblock Learning hand-eye coordination for robotic grasping with deep
  learning and large-scale data collection.
\newblock {\em arXiv preprint arXiv:1603.02199}, 2016.

\bibitem{pinto2016curious}
Lerrel Pinto, Dhiraj Gandhi, Yuanfeng Han, Yong-Lae Park, and Abhinav Gupta.
\newblock The curious robot: Learning visual representations via physical
  interactions.
\newblock {\em arXiv preprint arXiv:1604.01360}, 2016.

\bibitem{agrawal2016learning}
Pulkit Agrawal, Ashvin Nair, Pieter Abbeel, Jitendra Malik, and Sergey Levine.
\newblock Learning to poke by poking: Experiential learning of intuitive
  physics.
\newblock {\em arXiv preprint arXiv:1606.07419}, 2016.

\bibitem{bicchi2000robotic}
Antonio Bicchi and Vijay Kumar.
\newblock Robotic grasping and contact: A review.
\newblock In {\em ICRA}, pages 348--353. Citeseer, 2000.

\bibitem{bohg2014data}
Jeannette Bohg, Antonio Morales, Tamim Asfour, and Danica Kragic.
\newblock Data-driven grasp synthesis—a survey.
\newblock {\em Robotics, IEEE Transactions on}, 30(2):289--309, 2014.

\bibitem{lozano1989task}
Tom{\'a}s Lozano-P{\'e}rez, Joseph~L. Jones, Emmanuel Mazer, and Patrick~A.
  O'Donnell.
\newblock Task-level planning of pick-and-place robot motions.
\newblock {\em IEEE Computer}, 22(3):21--29, 1989.

\bibitem{nguyen1988constructing}
Van-Duc Nguyen.
\newblock Constructing force-closure grasps.
\newblock {\em IJRR}, 7(3):3--16, 1988.

\bibitem{miller2003automatic}
Andrew~T Miller, Steffen Knoop, Henrik~I Christensen, and Peter~K Allen.
\newblock Automatic grasp planning using shape primitives.
\newblock In {\em ICRA 2003}.

\bibitem{lenz2013deep}
Ian Lenz, Honglak Lee, and Ashutosh Saxena.
\newblock Deep learning for detecting robotic grasps.
\newblock {\em arXiv preprint arXiv:1301.3592}, 2013.

\bibitem{balorda1990reducing}
Zdravko Balorda.
\newblock Reducing uncertainty of objects by robot pushing.
\newblock In {\em ICRA}. IEEE, 1990.

\bibitem{balorda1993automatic}
Zdravko Balorda.
\newblock Automatic planning of robot pushing operations.
\newblock In {\em Robotics and Automation, 1993. Proceedings., 1993 IEEE
  International Conference on}, pages 732--737. IEEE, 1993.

\bibitem{lynch1996stable}
Kevin~M Lynch and Matthew~T Mason.
\newblock Stable pushing: Mechanics, controllability, and planning.
\newblock {\em The International Journal of Robotics Research}, 15(6):533--556,
  1996.

\bibitem{dogar2011framework}
Mehmet Dogar and Siddhartha Srinivasa.
\newblock A framework for push-grasping in clutter.
\newblock {\em Robotics: Science and systems VII}, 2011.

\bibitem{yun1993object}
Xiaoping Yun.
\newblock Object handling using two arms without grasping.
\newblock {\em The International journal of robotics research}, 12(1):99--106,
  1993.

\bibitem{zhou2016convex}
Jiaji Zhou, Robert Paolini, J~Andrew Bagnell, and Matthew~T Mason.
\newblock A convex polynomial force-motion model for planar sliding:
  Identification and application.
\newblock 2016.

\bibitem{katz2014perceiving}
Dov Katz, Arun Venkatraman, Moslem Kazemi, J~Andrew Bagnell, and Anthony
  Stentz.
\newblock Perceiving, learning, and exploiting object affordances for
  autonomous pile manipulation.
\newblock {\em Autonomous Robots}, 37(4):369--382, 2014.

\bibitem{fragkiadaki2015learning}
Katerina Fragkiadaki, Pulkit Agrawal, Sergey Levine, and Jitendra Malik.
\newblock Learning visual predictive models of physics for playing billiards.
\newblock {\em arXiv preprint arXiv:1511.07404}, 2015.

\bibitem{mottaghi2016happens}
Roozbeh Mottaghi, Mohammad Rastegari, Abhinav Gupta, and Ali Farhadi.
\newblock " what happens if..." learning to predict the effect of forces in
  images.
\newblock {\em arXiv preprint arXiv:1603.05600}, 2016.

\bibitem{schneider2009object}
Alexander Schneider, J{\"u}rgen Sturm, Cyrill Stachniss, Marco Reisert, Hans
  Burkhardt, and Wolfram Burgard.
\newblock Object identification with tactile sensors using bag-of-features.
\newblock In {\em Intelligent Robots and Systems, 2009. IROS 2009. IEEE/RSJ
  International Conference on}, pages 243--248. IEEE, 2009.

\bibitem{misra2016cross}
Ishan Misra, Abhinav Shrivastava, Abhinav Gupta, and Martial Hebert.
\newblock Cross-stitch networks for multi-task learning.
\newblock {\em CVPR}, 2016.

\bibitem{gkioxari2014r}
Georgia Gkioxari, Bharath Hariharan, Ross Girshick, and Jitendra Malik.
\newblock R-cnns for pose estimation and action detection.
\newblock {\em arXiv preprint arXiv:1406.5212}, 2014.

\bibitem{kokkinos2016ubernet}
Iasonas Kokkinos.
\newblock Ubernet: Training auniversal'convolutional neural network for low-,
  mid-, and high-level vision using diverse datasets and limited memory.
\newblock {\em arXiv preprint arXiv:1609.02132}, 2016.

\bibitem{ren2015faster}
Shaoqing Ren, Kaiming He, Ross Girshick, and Jian Sun.
\newblock Faster r-cnn: Towards real-time object detection with region proposal
  networks.
\newblock In {\em Advances in neural information processing systems}, pages
  91--99, 2015.

\bibitem{ioffe2015batch}
Sergey Ioffe and Christian Szegedy.
\newblock Batch normalization: Accelerating deep network training by reducing
  internal covariate shift.
\newblock {\em arXiv preprint arXiv:1502.03167}, 2015.

\bibitem{krizhevsky2012imagenet}
Alex Krizhevsky, Ilya Sutskever, and Geoffrey~E Hinton.
\newblock Imagenet classification with deep convolutional neural networks.
\newblock In {\em NIPS}, pages 1097--1105, 2012.

\bibitem{tieleman2012lecture}
Tijmen Tieleman and Geoffrey Hinton.
\newblock Lecture 6.5-rmsprop: Divide the gradient by a running average of its
  recent magnitude.
\newblock {\em COURSERA: Neural Networks for Machine Learning}, 4(2), 2012.

\end{thebibliography}

\end{document}